\documentclass[10pt,twocolumn,letterpaper]{article}

\usepackage[pagenumbers]{wacv} 

\usepackage{graphicx}
\usepackage{amsmath}
\usepackage{amssymb}
\usepackage{booktabs}
\usepackage{pifont}
\usepackage{adjustbox}
\usepackage[ruled, lined, longend]{algorithm2e}
\usepackage{graphicx}
\usepackage{float}
\usepackage{caption}
\usepackage{wrapfig}
\usepackage{xcolor}

\usepackage[pagebackref,breaklinks,colorlinks]{hyperref}

\usepackage[capitalize]{cleveref}
\crefname{section}{Sec.}{Secs.}
\Crefname{section}{Section}{Sections}
\Crefname{table}{Table}{Tables}
\crefname{table}{Tab.}{Tabs.}

\def\wacvPaperID{000} 
\def\confName{WACV}
\def\confYear{2025}

\begin{document}

\title{MONAS: Efficient Zero-Shot Neural Architecture Search for MCUs}

\author{Ye Qiao\\
University of California, Irvine\\
260 Aldrich Hall Irvine, CA 92697-1075\\
{\tt\small yeq6@uci.edu}
\and
Haocheng Xu\\
University of California, Irvine\\
260 Aldrich Hall Irvine, CA 92697-1075\\
{\tt\small haochx5@uci.edu}
\and
Yifan Zhang\\
University of California, Irvine\\
260 Aldrich Hall Irvine, CA 92697-1075\\
{\tt\small yifanz58@uci.edu}
\and
Sitao Huang\\
University of California, Irvine\\
260 Aldrich Hall Irvine, CA 92697-1075\\
{\tt\small sitaoh@uci.edu}
}
\maketitle

\begin{abstract}
Neural Architecture Search (NAS) has proven effective in discovering new Convolutional Neural Network (CNN) architectures, particularly for scenarios with well-defined accuracy optimization goals. However, previous approaches often involve time-consuming training on super networks or intensive architecture sampling and evaluations. Although various zero-cost proxies correlated with CNN model accuracy have been proposed for efficient architecture search without training, their lack of hardware consideration makes it challenging to target highly resource-constrained edge devices such as microcontroller units (MCUs). To address these challenges, we introduce MONAS, a novel hardware-aware zero-shot NAS framework specifically designed for MCUs in edge computing. MONAS incorporates hardware optimality considerations into the search process through our proposed MCU hardware latency estimation model. By combining this with specialized performance indicators (proxies), MONAS identifies optimal neural architectures without incurring heavy training and evaluation costs, optimizing for both hardware latency and accuracy under resource constraints. MONAS achieves up to a 1104$\times$ improvement in search efficiency over previous work targeting MCUs and can discover CNN models with over 3.23$\times$ faster inference on MCUs while maintaining similar accuracy compared to more general NAS approaches.
\end{abstract}

\section{Introduction}
Deep convolutional neural networks (CNN) have achieved incredible results in computer vision, speech recognition, object detection, and many other fields. Many manually designed CNNs aiming for high accuracy or computation efficiency have been proposed in computer vision, speech recognition, and other domains \cite{he2016deep,howard2017mobilenets, qiao2022two, kaeley2023support}. However, creating these network architectures requires significant amount of time, manual effort, expertise, and computing resources. Moreover, these fine-tuned models are often too large or compute-intensive to be deployed on edge devices like microcontroller units (MCUs) due to MCUs' limited memory, storage, and computation resources. To address this challenge, neural architecture search (NAS) is needed to automate CNN model design and explore specialized architectures within certain constraints \cite{lin2020mcunet}. Despite advancements in automated search techniques, most NAS approaches still suffer from time-consuming training and evaluation \cite{muNAS}.

To address the training time bottleneck in neural architecture search (NAS), zero-shot proxies have emerged as a promising technique \cite{white2023neural,krishnakumar2022nasbenchsuitezero,mellor2021neural,qiao2024micronas, qiao2024tg}. These proxies swiftly evaluate architectures with either no training at all or a single forward pass, conserving computational resources and accelerating NAS. However, existing works have largely overlooked resource-limited microcontroller units (MCUs), which dominate the low-power edge computing market, and have not incorporated real hardware awareness as feedback to optimize the architecture search.

To tackle this challenge, we introduce MONAS, an efficient NAS framework designed for MCUs. MONAS features a novel latency estimation model and combines multiple zero-shot proxies, each of which serves a unique purpose. By intelligently selecting efficient CNN architectures, MONAS enables edge devices to run CNN inference without prohibitive computational costs, streamlining the practical deployment of CNNs in dynamic edge computing environments. This work represents a significant advancement in efficient NAS solutions for discovering edge-targeted CNNs. The main contributions of this work are summarized as follows:
\begin{itemize}
  \item We propose MONAS, a novel zero-shot NAS framework that searches for the optimal CNN architectures for efficient MCU-based inference.
  \item We propose a hybrid objective function that combines a custom spectrum of the neural tangent kernel (NTK), the number of linear region counts, and hardware proxies, significantly improving NAS quality for MCUs. 
  \item We introduce a novel hardware inference latency estimation model that accurately predicts the real-world performance of MCUs. This model serves as both a zero-shot proxy and a hardware constraint during the search process.
  \item We propose a novel hardware-aware pruning-based search algorithm to improve the search efficiency under resource constraints. 
  \item Our analysis results of trainless proxies on NAS-Bench-201 space show that MONAS could achieve great improvement in search efficiency over previous work targeting similar MCUs and can discover CNN models with over 3.23$\times$ faster inference on MCUs while maintaining similar accuracy.
\end{itemize}

 \begin{figure*}[ht]
    \centering 
    \includegraphics[width=1\textwidth]{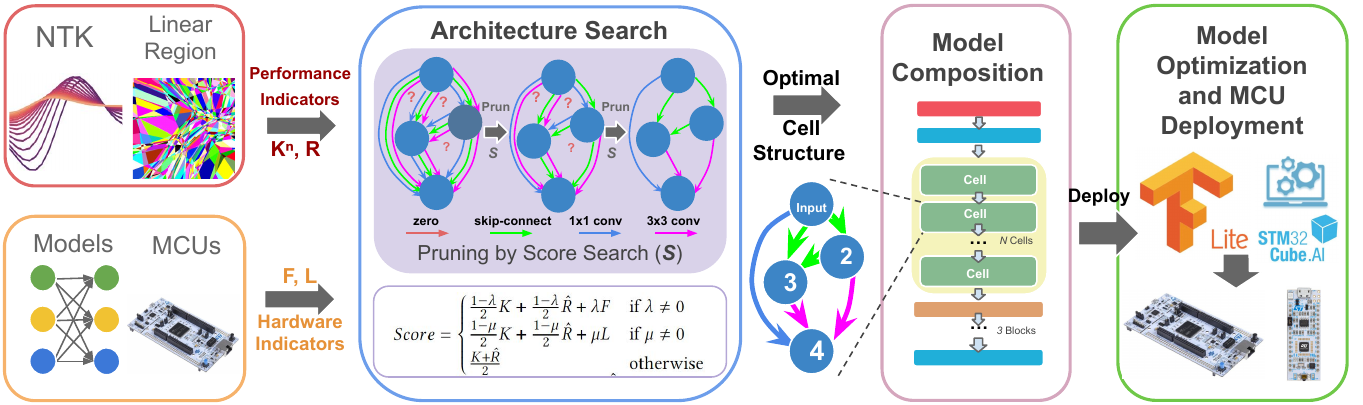}
    \caption{Overview of Proposed MONAS Workflow}
    \label{fig:framwork}
\end{figure*}


\section{Background and Related Works}

Efficient deep learning for edge devices is a challenging problem for both AI and hardware communities. The deep multi-layer nature of DNN models prevents them from being deployed directly to resource-constrained edge devices like MCUs. 
NAS and zero-cost proxies are proposed to discover new efficient model architectures specialized for edge environments \cite{lin2020mcunet}.

\subsection{Neural Architecture Search (NAS)}
The traditional NAS approach comprises two primary components, the architecture generator and the accuracy predictor. The generator is responsible for proposing potential DNN architectures and the predictor evaluates the accuracy of these architectures. Notably, reinforcement learning (RL) \cite{tan2019mnasnet} and evolutionary algorithm (EA) \cite{real2019regularized} have emerged as the most popular approaches for generating candidate architectures. However, both RL and EA require lengthy DNN training to get the evaluation results due to the lack of accuracy predictor. To address this problem, 
one-shot methods, for example, ProxylessNAS \cite{cai2019proxylessnas} and Once-for-all \cite{cai2020onceforall}, were proposed to further reduce the training and evaluation cost by sampling sub-networks (subnets) from a pre-trained super network (supernet). MCUNet \cite{lin2020mcunet} employs a two-stage NAS approach, where the first stage narrows the search space based on hardware constraints, while the following one-shot evaluation stage integrates latency optimization through the TinyEngine library. Noteworthy is that MCUNet did not consider the inference latency on the target hardware and DNN's number of floating-point operations (FLOPs) during NAS as shown in the table \ref{table1}. Furthermore, MCUNet only considers MobileNetV2 \cite{sandler2019mobilenetv2} like architectures as its foundational search space, distinct from more general cell-based spaces like DARTS \cite{liu2018darts} and NAS-Bench-201 \cite{dong2020nasbench201}, which greatly limits its generality. This limitation prompts careful re-consideration of MCUNet's trade-offs and advantages within the broader NAS landscape. Our proposed MONAS considers the NAS problem in general cell-based search spaces for maximum generality. 

\subsection{Zero-Shot Proxies}
\renewcommand{\arraystretch}{1.1}
    

\begin{table*}[]
\begin{adjustbox}{width=\textwidth}
\begin{tabular}{c|cccc}
\hline
Method       & Search Space                                                                                                      & Hardware Awareness & Search Strategy                                                                      & Search Time \\ \hline
MCUNet\cite{lin2020mcunet}       & \begin{tabular}[c]{@{}c@{}}MobileNetV2-like\cite{MnasNet} \\ (Input resolution and the width multiplier)\end{tabular}       & \ding{55}               & Two Stages Co-search                                                                 & Long        \\
$\mu$NAS\cite{muNAS}         & \begin{tabular}[c]{@{}c@{}}Custom Search Space \\ ($N$ conv blocks, $K$ channels, and others)\end{tabular}            & \ding{55}                 & \begin{tabular}[c]{@{}c@{}}Aging Evolution with\\ Bayesian Optimisation\end{tabular} & Long        \\
\textbf{MONAS (Ours)} & \begin{tabular}[c]{@{}c@{}}Cell-based Search Space\\ (e.g. NASBench201\cite{dong2020nasbench201}, DARTS\cite{liu2018darts})\end{tabular} & \textbf{\ding{51}}                & \begin{tabular}[c]{@{}c@{}}Hardware Aware\\ Prune-based Search\end{tabular}          & \textbf{Short}       \\ \hline
\end{tabular}
\end{adjustbox}
    \footnotesize{*Our method can extend to DARTS\cite{liu2018darts} search space. However, due to the hardware constraints of MCU, most operators/layers found in DARTS search space can not be supported. }
    \caption{Comparison with other NAS targeting MCU devices}
    \label{table1}

\end{table*}

Conventional NAS algorithms find decent neural architectures but these architectures are likely impractical for resource-constrained edge computing devices due to high computational costs. Recent works explore the concept of zero-shot proxies for DNN accuracy prediction to enable efficient NAS without significant computational burdens. 

One commonly used accuracy indicator is the expressivity of a neural network, which relates to the complexity of the function it can represent. Especially, the number of linear regions has been widely used as an expressivity indicator of the neural network. Several recent works, such as TE-NAS \cite{chen2021neural}, ZEN-NAS\cite{lin2021zennas}, and NASWOT \cite{mellor2021neural}, approximate the behavior of neural networks by considering the expressivity of ReLU network and the number of linear regions they can separate.

Kernel methods have been employed to approximate the convergence and generalization ability of networks without training. For instance, TE-NAS \cite{chen2021neural} formulates neural networks as a Gaussian Process \cite{lee2019wide,yang2020scaling} and analyzes randomly initialized architectures by the spectrum of the neural tangent kernel (NTK) \cite{jacot2020neural,xiao2020disentangling} and the number of linear regions in the input space. Zico \cite{li2023zico} extends this kernel-based analysis to reveal the relationships between gradient properties, training convergence, and the generalization capacity of neural networks. Similarly, gradients with respect to the parameters of neural networks are proven to approximate the Taylor expansion of deep neural networks \cite{lee2019snip,wang2020picking,abdelfattah2021zerocost}. SNIP \cite{lee2019snip} introduced a saliency metric computed at initialization using a single minibatch of data. Grasp \cite{wang2020picking} improved upon SNIP by approximating the change in gradient norm instead of loss. Synflow \cite{abdelfattah2021zerocost} generalized previous approaches and proposed a modified version that computes a loss as the product of all parameters in the network, thus requiring no data. TG-NAS \cite{qiao2024tg} leverages a universal model-based predictor via a transformer operator embedding generator. Additionally, gradients with respect to feature maps combined with the number of linear regions have been used in Zen-NAS \cite{lin2021zennas}.

\subsection{Search Space for NAS}
In addition to reducing the cost of accuracy evaluation through introducing zero-cost proxy, the vastness of the search space remains a significant challenge in evaluating the performance of NAS algorithms. The search space in NAS refers to the immense number of possible neural network architectures that can be generated and evaluated. Each architecture may possess unique combinations of layers, connections, activation functions, and other design choices. Exhaustively evaluating the performance of every potential architecture within the search space would require a prohibitive amount of computational resources and time. However, recent research \cite{ying2019nasbench101,dong2020nasbench201,zela2022surrogate,li2021hwnasbenchhardwareaware} has made significant progress in addressing this issue by introducing benchmarks that offer tractable NAS search spaces and accompanying metadata for the training of networks within those search spaces. These benchmark datasets provide predefined and manageable search spaces that capture a representative subset of possible neural network architectures. By constraining the search space, researchers can systematically explore and evaluate a diverse range of architectures without the need for exhaustive exploration. Additionally, the availability of metadata, such as training settings and performance metrics, enables fair and consistent comparisons between different NAS algorithms. In this work, we choose NAS-Bench-201 \cite{dong2020nasbench201} as our search space, which includes all possible architectures generated by 4 nodes and 5 associated operation options, which results in 15,625 neural cell architecture candidates.

\subsection{Hardware Constraints and Proxies}
For effective consideration of hardware constraints on MCUs, NAS algorithms must incorporate hardware indicators. While some NAS works have used FLOPs as a proxy metric for latency \cite{mei2020atomnas}, relying solely on this metric may overlook crucial aspects like instruction-level scheduling and parallelism. MCUs hardware and software specifications are highly diverse so we need proxies to represent them. Considering factors beyond FLOPs is essential to capturing the unique characteristics of MCU hardware and software stack for efficient inference at the edge. The recent latency predictor model \cite{akhauri2024latency} is designed for larger devices like CPUs, GPUs, and ASIC accelerators rather than MCUs. 


\section{MONAS Framework}

To eliminate the need for training and evaluation during the architecture search process, we propose and combine key indicators that capture the trainability \cite{lee2019wide,chen2021neural,xiao2020disentangling,jacot2020neural}, expressivity \cite{lin2021zennas,mellor2021neural}, and hardware performance of neural networks. We incorporate multiple zero-cost proxies and generalize the search to discover the optimal cell structure for MCUs. A cell-based search space defines each DNN architecture as a directed acyclic graph (DAG), with nodes representing feature maps and edges representing operations, as shown in Figure \ref{fig:framwork}.


\subsection{Performance Indicators}
\subsubsection{Spectrum of Neural Tangent Kernel}
The neural tangent kernel (NTK) is a mathematical construct used to analyze the behavior and properties of neural networks \cite{jacot2020neural,xiao2020disentangling,lee2019wide}. The trainability of a neural network, which refers to its convergence and generalization ability, is a crucial aspect of zero-shot NAS. The finite-width NTK, denoted as $\hat{\Theta}(x,x')=J(x)J(x')^\top$, where $J(*)$ is the Jacobian and $x$ and $x'$ are the data points, captures the behavior of a neural network at initialization, thus analyzing the spectrum of the infinite NTK allows us to formulate the necessary conditions for trainability and generalization \cite{xiao2020disentangling}. 

\begin{figure}[ht]

    \centering
    \begin{subfigure}[b]{.48\columnwidth}
        \centering
        \includegraphics[width=\linewidth]{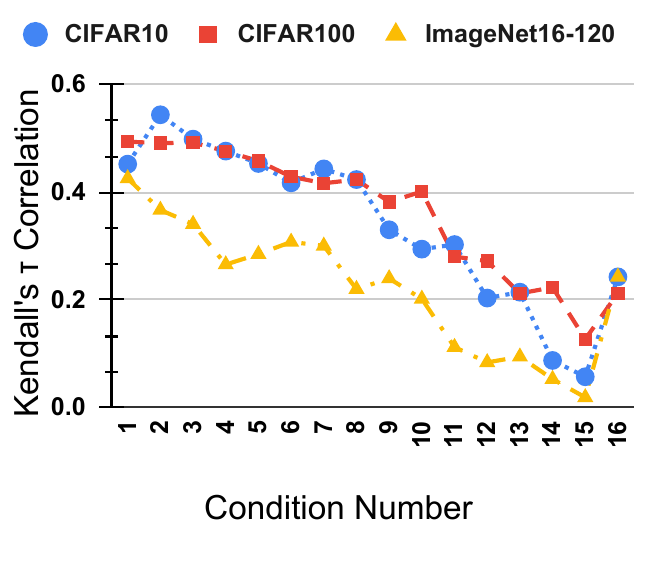}
 
        \caption{Kendell's $\tau$ vs. $K_i$}
        \label{ntkvst}
    \end{subfigure}
    \hfill
    \begin{subfigure}[b]{.48\columnwidth}
        \centering
        \includegraphics[width=\linewidth]{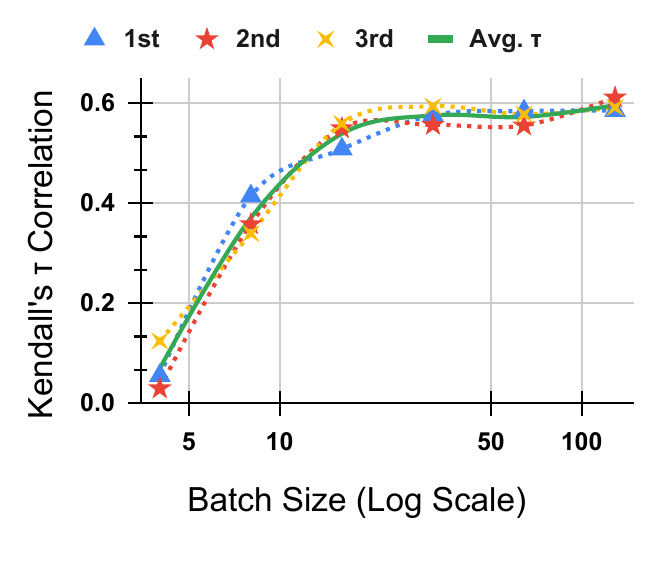}
      
        \caption{Kendall-$\tau$ vs. Batch Size}
        \label{batchvst}
    \end{subfigure}
  
    \caption{Kendall-$\tau$ vs Batch Size, Condition Number $K_i$ }
    \label{fig:tau}
\end{figure}

Once we decomposed the NTK, the training dynamic can be solved as
\begin{equation}
    \mu_t(X_{train}) = (I-e^{-\eta\lambda_it}) \cdot Y_{train,t}
    \label{eq2}
\end{equation}
\noindent where $\lambda_i$ are the eigenvalues of $\hat{\Theta}_{train}$ and $\lambda_1$ to $\lambda_n$ are in descending order. As suggested by \cite{xiao2020disentangling}, there are four different indicators of the spectrum associated with the convergence of neural networks. One of them is conditional number $K = \frac{\lambda_1}{\lambda_n}$. We conducted the experiments three times and reported the average Kendall-$\tau$ correlation \cite{kendall1938new} versus different variation forms of the condition number $K_i=\frac{\lambda_1}{\lambda_{n-i}}$. Figure \ref{ntkvst} illustrates the trends of Kendall-$\tau$ correlation with the different condition number $K_i$. In our search, we empirically choose $K_2$ since it results in the best average correlation score.

Since the condition number of NTK is calculated using a single mini-batch of the data, the batch size will affect the consistency of the NTK spectrum, ultimately influencing the search results. To further understand this effect, we investigate the relationship between Kendall-$\tau$ correlation and logarithmic scales of batch size. From Figure \ref{batchvst}, we observe that the optimal batch size for balancing the performance indicator and the search cost lies between 16 and 32 across three datasets. Increasing the batch size beyond 32 up to 128 does not significantly alter the Kendall-$\tau$ correlation, but it exponentially increases the search cost. In our experiments, we fix the batch size to 32 for optimal search cost and results.


\subsubsection{Linear Region Count} In our study, we examine the expressivity of a vanilla CNN, where each layer consists of a single convolutional operator followed by the ReLU activation function. Since ReLU is a piecewise linear function, the input space of the network can be separated into distinct linear regions (LR) \cite{xiong2020number}. Every LR is associated with a set of affine parameters, and the function represented by the network is affine when restricted to each LR. Thus the expressivity of a neural network can be indicated by the number of linear regions it can separate. To quantify the expressivity, we define the LR count of a neural network as


\begin{equation}
    \hat{R}\approx E_{\theta} \cdot R_{\theta}
\end{equation}
where $R_{\theta}$ denotes the number of linear regions at $\theta$ and we calculate the average of LR counts as an approximation to its expectation $E_{\theta}$. \cite{xiong2020number,chen2021neural} demonstrates that $\hat{R}$ and $K$ are positively and negatively correlated to the model accuracy respectively.


\subsection{Hardware Indicators}
The hardware-aware aspect of the search process necessitates meticulous attention to platform-specific constraints. In this study, our emphasis is placed on low-power microcontroller units (MCUs). Considering the severely constrained hardware resources of these devices, it becomes imperative to effectively manage computation costs, processing latency, and memory utilization. Among these factors, we assign higher importance to inference latency as it directly corresponds to real-world processing time. Consequently, we incorporate both number of floating-point operations (FLOPs) estimation, denoted as $F$, and hardware latency modeling, denoted as $L$, into the architecture search process for the sampled models. Our approach provides tunable weight factors that enable fine-grained control over the contributions of $F$ and $L$ during the search.

\begin{figure}
 
    \centering
    \includegraphics[width=\linewidth]{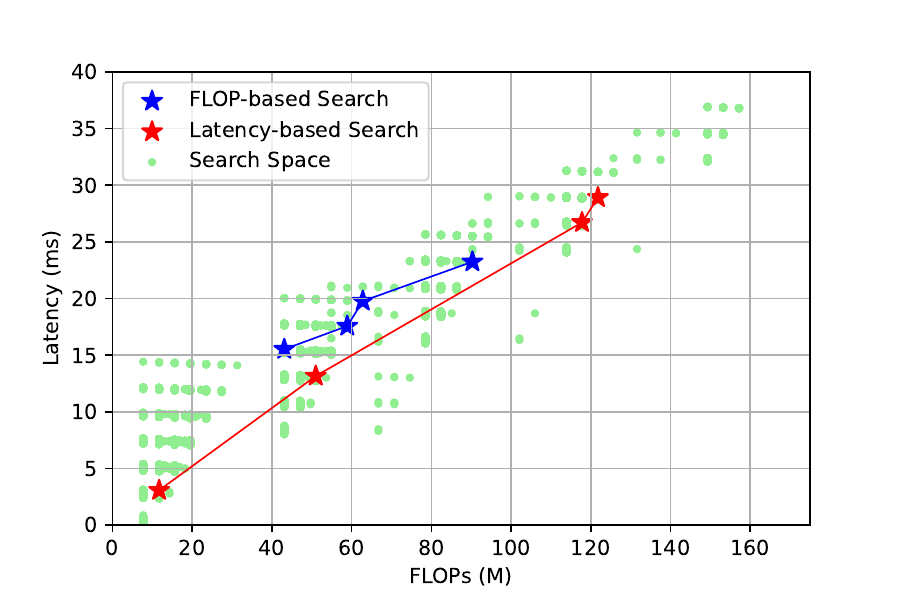}

    \caption{Model FLOPs vs. Latency Spectrum on MCU}
    \label{fig:FLOP_Latency_all}

\end{figure}

\subsubsection{Number of Floating-Point Operations (FLOPs)}


FLOPs count is a significant indicator of deep learning model complexity, reflecting computing time in the target environment regardless of hardware specialization or optimization. Our FLOPs estimation considers operations in convolution, batch normalization, pooling, and fully connected layers. While FLOPs count is usually positively correlated with model accuracy, it is important to note that FLOPs alone does not represent absolute accuracy, robustness, or real-world hardware performance, given the redundancy and topological differences in convolutional neural networks. Therefore, we introduce estimated model inference latency as an additional hardware indicator to further assess and compare different DNN models.

\subsubsection{Estimated Inference Latency}

Model inference latency on MCUs is influenced by factors such as software and hardware scheduling, caching, and parallelism. However, when using the number of multiply-accumulate operations (MACs) as a proxy for model latency as in $\mu$NAS \cite{muNAS}, the intricate details of operations in highly optimized neural network inference libraries for MCUs are neglected. As shown in Figure \ref{fig:FLOP_Latency_all}, the latency observed in NAS-Bench-201 with STM32AI Deep Learning Library on STM32F7 MCU is not directly proportional to reported FLOPs due to diverse implementations and optimizations for different deep learning (DL) operators.
\RestyleAlgo{ruled}
\SetKwComment{Comment}{/* }{ */}
\begin{algorithm}[hbt!]
\caption{Inference Latency Estimation}\label{alg:1}
\KwData{$Searched\_cells, latency\_table, const\_overhead$}
\KwResult{Latency estimation $L$}
 $L \gets 0$ \;
 \For{$idx\_cell, cell$ \textbf{in} $Searched\_cells$}{
    masks $\gets$ cell.get\_operations\_by\_edge $>$ 0\, ? \,1 : 0\;
    \For{$edge\_mask$ \textbf{in} $masks$}{
    latency\_edge $\gets$ latency\_table(edge\_mask)\;
    $L \gets L + $ latency\_edge\;
    }
}
\Return{$L$ + const\_overhead}
\end{algorithm}

\begin{figure}[!ht]
    \centering
    \begin{subfigure}[b]{.55\columnwidth}
        \centering
        \includegraphics[width=\linewidth]{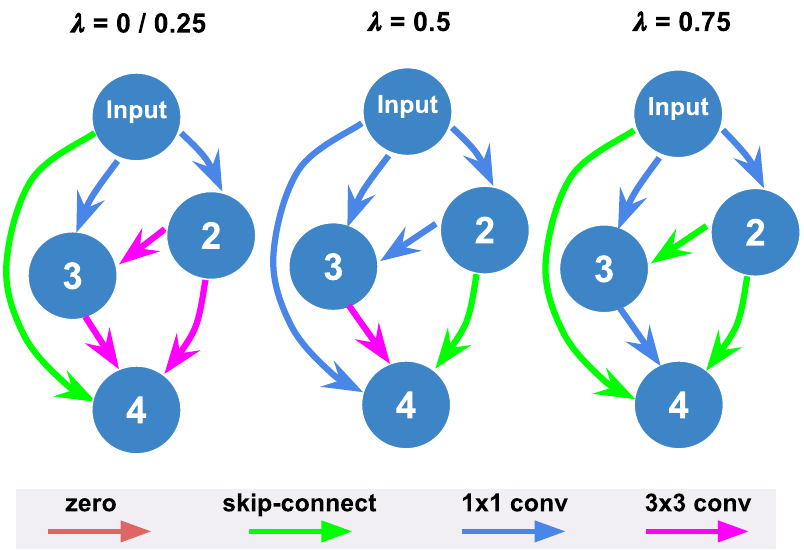}
        \caption{Searched Cell Architectures}
        \label{fig:cells}
    \end{subfigure}
    \hfill
    \begin{subfigure}[b]{.43\columnwidth}
        \centering
        \includegraphics[width=\linewidth]{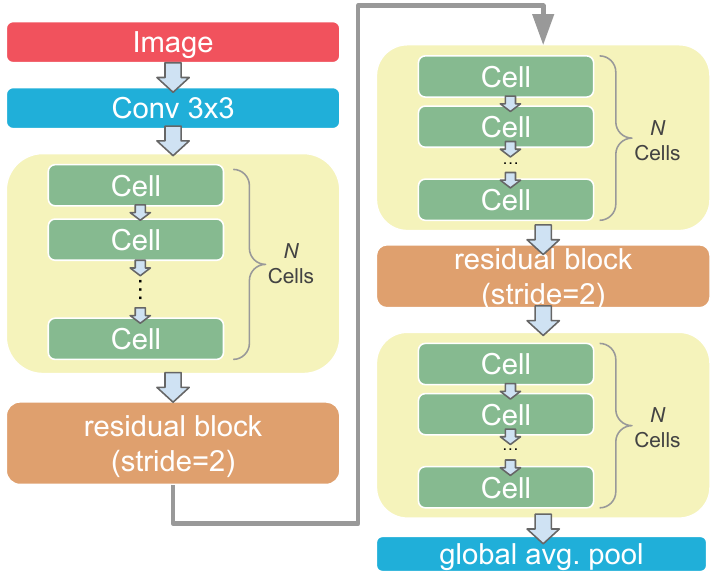}
        \caption{\centering The macro skeleton of each architecture candidate}
        \label{fig:macro_skeleton}
    \end{subfigure}
    \caption{Architecture of Searched Cell and Network}
    \label{fig:arch}
\end{figure}

To address this, we propose and build a custom latency estimator to accurately model inference latency on any target MCUs based on provided cell structures. We build the estimator by profiling each operation individually within the search space and generating a reference lookup table. Specific details of the secondary stage of the model structure, including the number of cells and input/output channels for each cell, are gathered. Predefined search spaces with their fixed secondary stage provide meaningful comparisons. For example, in NAS-Bench-201 space, three searched cells are put in the stack, supplemented by additional ResNet blocks interspersed between them and an initial first convolution layer (Figure \ref{fig:macro_skeleton}). Finally, constant hardware latency overhead is profiled and incorporated into the overall latency estimation (see Algorithm \ref{alg:1} for details). It is worth noting that the increase in searching overhead is negligible, given the lightweight nature of our proposed analytical latency model. As our primary focus is the relative ranking of the latency (instead of absolute latency) to guide architecture search, our proposed approach is portable across different devices. We verify the portability of this modeling approach by searching for large and small MCUs, and it achieves  good DNN inference results as shown in the later result section. We believe our approach can also easily adapt to other types of edge devices such as Raspberry Pi.

\subsection{Hardware-Aware Pruning-based Architecture Search}


Traditional NAS methods can be classified into two primary types. The first type involves applying reinforcement learning or evolutionary algorithms in conjunction with a sampling-based method to explore a discrete search space. These methods sample and evaluate architectures to optimize the searching agent's reward. The second type adopts an architecture representation relaxation, enabling a more efficient search process using gradient descent. Here, the architecture is represented continuously, allowing for the application of gradient-based optimization techniques to search for optimal architectures \cite{liu2018darts}. However, their effectiveness is limited, particularly when dealing with intricate cell-based search spaces. Consider a network comprising interconnected cells in a directed acyclic graph structure. Each cell contains $E$ edges, with each edge selecting a single operator from a set of $\lvert O \rvert$ potential candidates. Consequently, there are $|O|^E$ distinct cells, and during the sampling-based search process, $\alpha|O|^E$ networks must be sampled. The parameter $\alpha$ represents the sampling process efficiency, where a smaller value can help to find superior architectures at a faster rate. Nevertheless, the computational burden of sampling-based methods remains dependent on the scale of the search space, \emph{i.e.}, $|O|^E$.


To enhance efficiency and expedite the search process, we propose a pruning-by-score mechanism, inspired by Lee et al \cite{lee2018snip}. This approach reduces the search cost from $\alpha|O|^E$ to $|O|\cdot E$. Initially, the search begins with a super-network $N_{init}$ encompassing all possible operators for each edge. In the outer loop, we iteratively prune one or multiple operators on each edge, depending on the specified configuration. In cases with strict time constraints, simultaneous pruning of multiple edges may be advantageous to reduce search time, even if it leads to potentially inferior final results. The outer loop concludes when the current supernet $N_t$ transforms into a single-path network, representing the final searched architecture.

\RestyleAlgo{ruled}
\SetKwComment{Comment}{/* }{ */}

During the inner loop, we assess the impact of pruning each individual operator on metrics such as $K_N$, $\hat{R}$, $F$, and $L$. The operator's importance is evaluated using a $Score$ function as shown in Equation \ref{eq1}, where $\lambda$ and $\mu$ are mutually exclusive parameters from the range (0,1), with only one hardware indicator employed at a time. We conducted experiments with various combinations of these indicators, but their individual application yielded superior results.

\SetAlFnt{\small}
\begin{algorithm}[htb]

\caption{MONAS Search Algorithm}\label{alg:2}
\KwData{$M_0$ supernet, $E$ edges in each cell, every edge have $O$ operators}
\While{$M_t$ is not a single path network}{
 \For{$Cell\_type$, $ct$ \textbf{in} $M_t$}{
    \For{edge $E_i$,  operator $O_j$ \textbf{in} $ct$, $E_i$}{
                $\Delta (K,R,F,L)_{t, e_i, o_j} \gets (K,R,F,L)_{t, e_i, o_j} - (K,R,F,L)_{t, e_i \backslash o_j}$ \;
    }
    $Rank_K$ $\gets$ index of $o_j$ in descending of $\Delta K_t$\;
    $Rank_F$ $\gets$ index of $o_j$ in descending of $\Delta F_t$\;
    $Rank_L \gets$ index of $o_j$ in descending of $\Delta L_t$\;
    $Rank_R \gets$ index of $o_j$ in ascending of $\Delta R_t$\;
    $S \gets$ Score($Rank_K$, $Rank_R$, $Rank_F$, $Rank_L$)\;
    
 }

 \For{$e_i$ in $E$}{
    $o_p \gets argmin\{S(o_p)\:  ,o_p \in e_i\}$ \;
    $M_{t+1} \gets M_t - o_p$\;
 }
 $t \gets t + 1$\;
}
\Return{single path network searching result $M_t$}\;

\end{algorithm}

Subsequently, we rank the available operators based on their scores and eliminate the operator with the lowest importance on each edge. In the innermost loop, we compute the average value for each indicator through a specified number of repetitions to ensure rank consistency, especially for NTK, which exhibits a wide range of values. Detailed steps are provided in Algorithm \ref{alg:2}.

\begin{equation}
    Score =
    \begin{cases}
      \frac{1-\lambda}{2}K + \frac{1-\lambda}{2}\hat{R} + \lambda F & \text{if $\lambda \neq 0$}\\
      \frac{1-\mu}{2} K + \frac{1-\mu}{2} \hat{R} + \mu L & \text{if $\mu \neq 0$} \\
      \frac{K + \hat{R}}{2} & \text{otherwise}
    \end{cases}
    \label{eq1}
\end{equation}

The four indicator measurements $K$, $\hat{R}$, $F$, and $L$ are combined by weighted summation of their relative ranks with a given weight. This is to reduce the imbalance effect caused by the significantly different value ranges of the three measurements. The weight of $K$ and $\hat{R}$ is set to equal based on empirical results \cite{chen2021neural}. The weight of $F$ and $L$ is constrained by hardware resources and can be easily tuned for different target devices.


\begin{table*}[ht]

\caption{Results of NAS-Discovered DNNs Running on CIFAR-10, CIFAR-100 and ImageNet16-120, with Small and Large MCUs}

\centering
\begin{adjustbox}{width=\textwidth}
\begin{tabular}{lcccccccc}
\hline
NAS Frameworks &
  FLOPs (M) &
  Params (M) &
  \begin{tabular}[c]{@{}c@{}} Inference\\Latency (ms)\end{tabular} &
  \begin{tabular}[c]{@{}c@{}}Latency\\Speedup\end{tabular} &
  \begin{tabular}[c]{@{}c@{}}Avg. Search Time\\ (GPU Hours)\end{tabular} &
  \begin{tabular}[c]{@{}c@{}}CIFAR10\\Accuracy (\%) \end{tabular} &
  \begin{tabular}[c]{@{}c@{}}CIFAR100\\Accuracy (\%) \end{tabular} &
   \begin{tabular}[c]{@{}c@{}}ImageNet16-120\\Accuracy (\%)\end{tabular}\\ \hline
$\mu$NAS\cite{muNAS}   & -              & 0.014        &-  & -              & 552            & 86.49          & -   & -           \\
DARTS\cite{liu2018darts}    & 82.49        & 0.587     &20.80         & 2.04$\times$              & 3.02           & 88.32          & 67.78         &34.6 \\
GDAS\cite{dong2019searching}   & 117.88     & 0.83     &28.83        & 1.47$\times$              & 8.03           & 93.36          & 69.64   &38.87      \\
KNAS\cite{knas}        & 153.27            & 1.073     &36.85        & 1.15$\times$              & 2.44           & 93.43          & 71.05    &45.05      \\
NASWOT\cite{mellor2021neural}  & 94.29     & 0.671     &25.45        & 1.67$\times$              & 0.09           & 92.96          & 69.7     &44.47     \\
TG-NAS\cite{qiao2024tg}  & 113.95         & 0.802  &42.48        & 1.74$\times$             & 0.01           & 93.75          & 70.64    &44.97     \\
ZiCo\cite{li2023zico}  & 153.27         & 1.073  &34.45        & 1.23$\times$             & 0.40           & 94.00          & 71.10    &41.80     \\
TE-NAS\cite{chen2021neural}  & 188.66         & 1.317  &42.48        & 1             & 0.43           & 93.78          & 70.44    &41.40     \\
\hline
MONAS Baseline (Large MCU)  & 149.34  & 1.045 & 32.30 & 1.32$\times$ & 0.50 & \textbf{94.08} & 72.01 & 45.62 \\ 
MONAS (Large MCU) {$\lambda$}=0.7     & \textbf{58.90}  & 0.428 &17.56 & 2.42$\times$ & 0.50 & 92.31 & 71.12 &45.07 \\
MONAS (Large MCU) {$\mu$}=0.5/0.6     & 121.82  & 0.858 &28.91 & 1.47$\times$ & 0.50 & \textbf{93.88} & \textbf{73.16} &\textbf{45.67} \\
MONAS (Large MCU) {$\mu$}=0.7     & 117.88  & 0.83   &26.71 & 1.59$\times$ & 0.50 & 93.72 & 70.67 &44.22\\
MONAS (Large MCU) {$\mu$}=0.8    & \textbf{51.04} & \textbf{0.372}   &\textbf{13.15} & \textbf{3.23$\times$}  & 0.50 & 92.67 &68.94 &43.33 \\ \hline
MONAS (Small MCU)\textbf{*} {$\mu$}=[0, 0.7]   & 153.27 & 1.073   &154.70 & -  & 0.50 & \textbf{94.37} &70.09 &\textbf{46.33} \\
MONAS (Small MCU)\textbf{*} {$\mu$}=0.8    & 47.10 & 0.344   &48.94 & -  & 0.50 & 92.21 &67.41 &40.32 \\
MONAS (Small MCU)\textbf{*} {$\mu$}=0.85    & 11.72 & 0.101   &23.11 & -  & 0.50 & 89.72 &63.79 &33.58 \\
MONAS (Small MCU)\textbf{*} {$\mu$}=0.88    & 11.72 & 0.101   &\textbf{14.62} & -  & 0.50 & 88.51 &58.83 &30.56 \\ \hline

\end{tabular}%
\end{adjustbox}
    \footnotesize{* Searching for and performing inference on the smaller STM32L43 MCU operating at 80MHz.}




\label{table2}
\end{table*}



\section{Experiment and Results Analysis}
\subsection{Experimental Setup} 
We evaluate MONAS on NAS-Bench-201 \cite{dong2020nasbench201}, which is a cell-based NAS benchmark widely used for performance evaluation for zero-shot NAS. The macro skeleton of the approach, as illustrated in Figure \ref{fig:macro_skeleton}, comprises three stacks of cells. Each cell is composed of four nodes and offers five operator options: \textit{none (zero), skip-connection, conv1×1, conv3×3, avepooling}. The search space contains in total of 15,625 architectures. Each architecture is trained on three different image datasets for 200 epochs: CIFAR-10, CIFAR-100, and ImageNet16-120. The NAS process runs on an NVIDIA RTX 3090. The resulting DNN from the search is compiled by X-CUBE-AI compiler and deployed on two MCUs, a \textbf{large} MCU, STM32F746ZG board (\textit{ARM Cortex-M7 core@216MHz, 6-stage dual-issue, 320KB SRAM, 1MB Flash}), and a \textbf{small} MCU, STM32L43 ultra-low power board (\textit{ARM Cortex-M4 core@80MHz, 3-stage single-issue, 64KB SRAM, 128KB Flash}). 

\subsection{Experimental Results}

\subsubsection{Search Results Analysis} 
\begin{figure}[htbp]
    \centering
    \begin{subfigure}[b]{\linewidth}
        \centering
    \includegraphics[width=\linewidth]{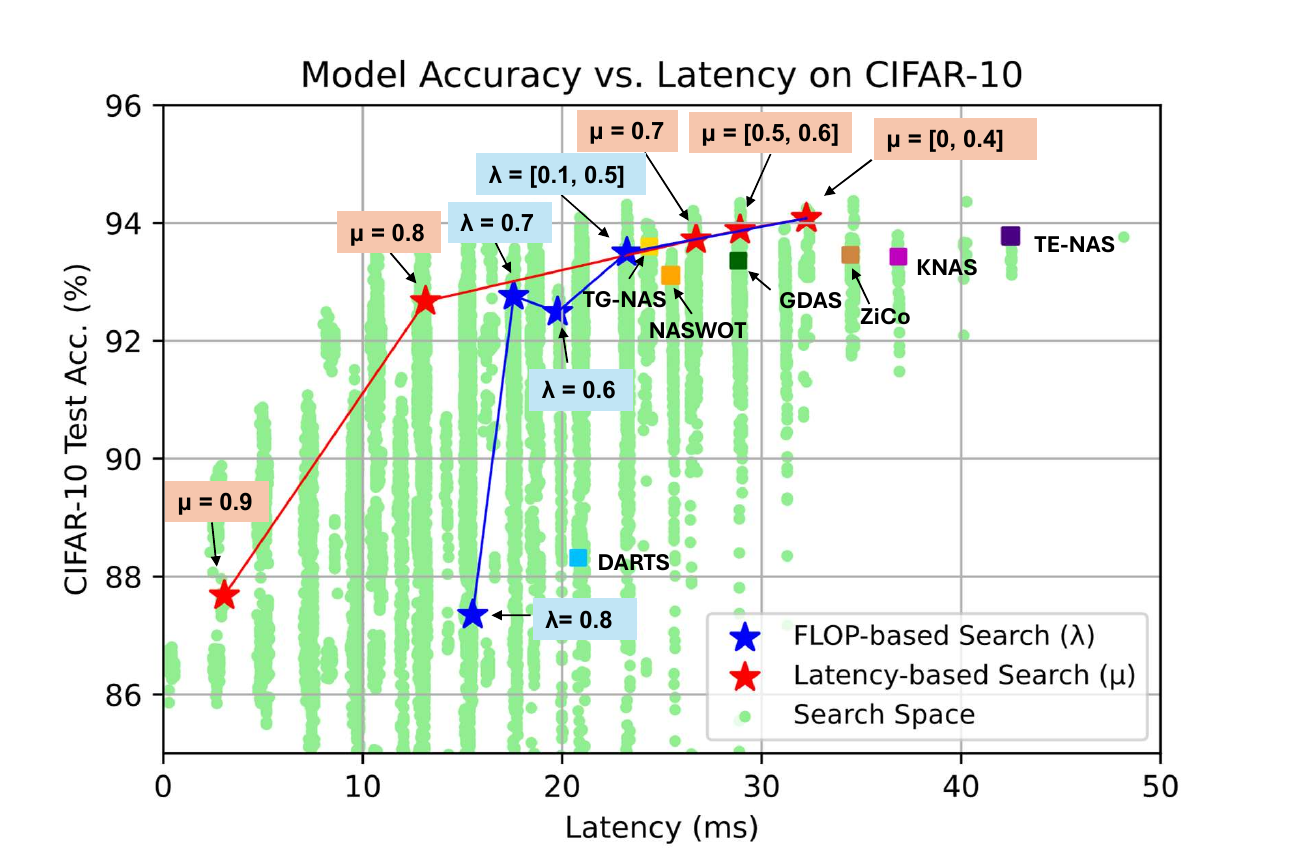}
    \end{subfigure}
    \vfill
    \begin{subfigure}[b]{\linewidth}
        \centering
    \includegraphics[width=\linewidth]{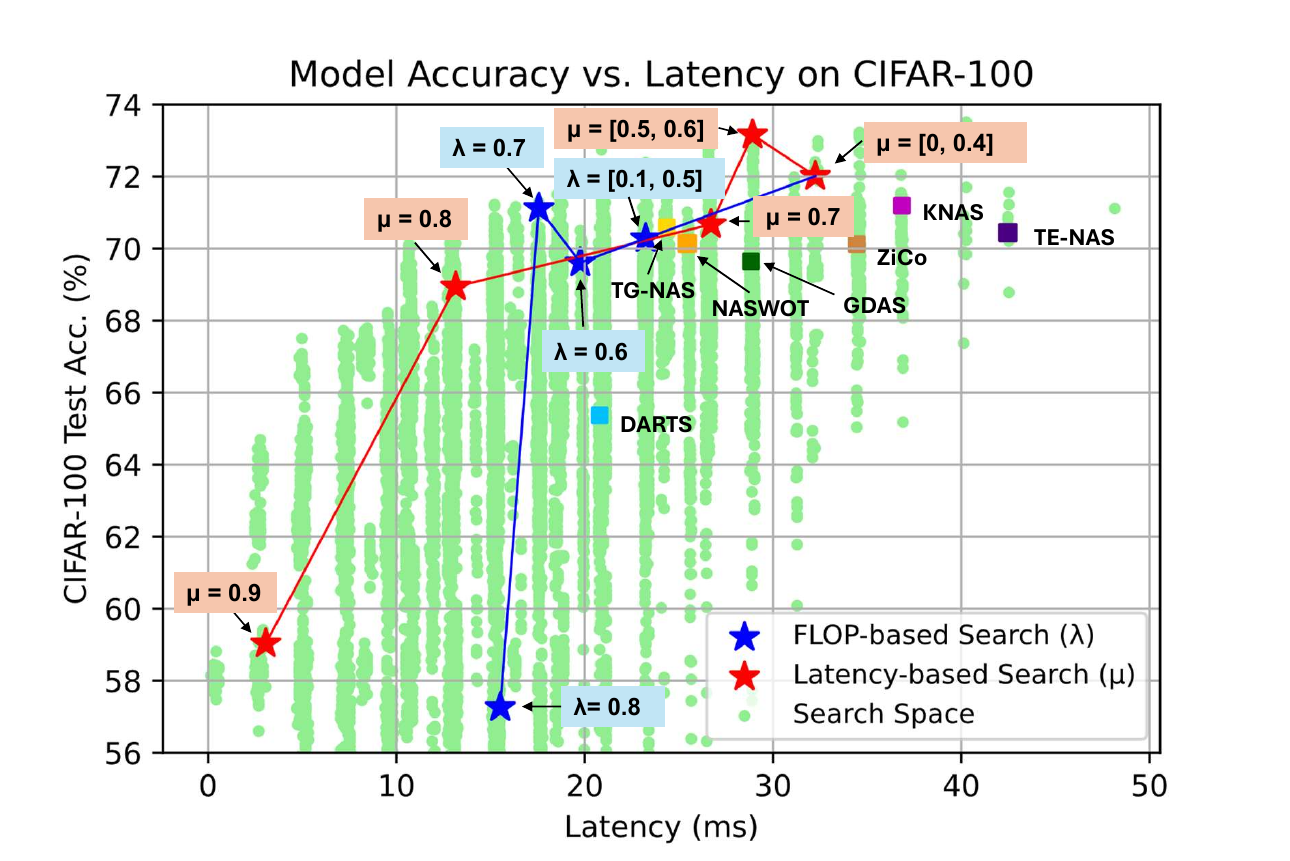}
    \end{subfigure}
    \vfill
    \begin{subfigure}[b]{\linewidth}
        \centering
    \includegraphics[width=\linewidth]{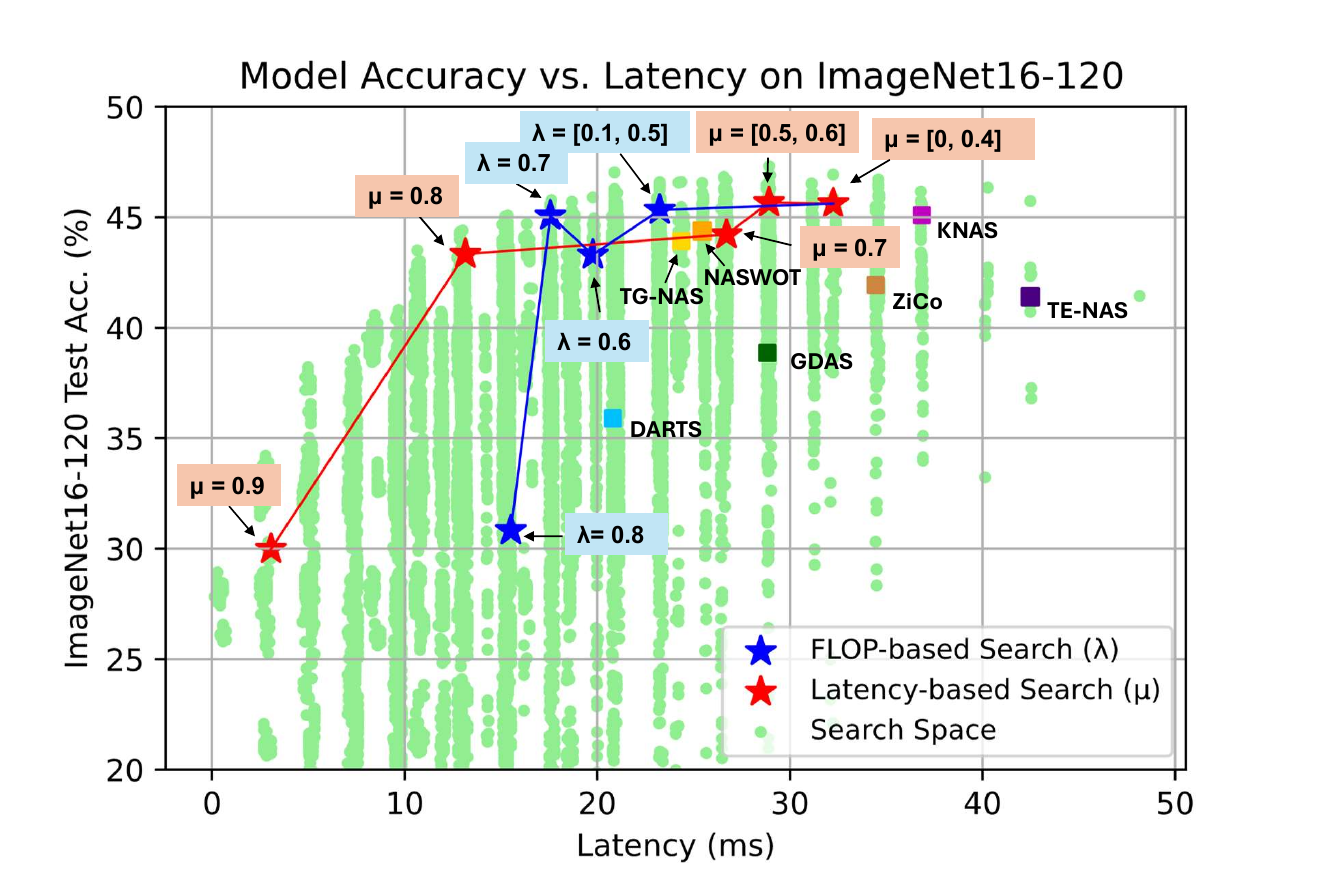}
    \end{subfigure}
\caption{Evaluation Results on STM32F7 MCU}
\label{STM32F7}
\end{figure}

In the search, the weight for FLOPs and latency is denoted as $\lambda$ and $\mu$ respectively. Figure \ref{fig:cells} shows searched cells with different $\lambda's$ to demonstrate distinct hardware constraints. We conduct the evaluation by varying weight values for FLOPs and latency estimation, and the search was performed on three representative datasets: CIFAR-10, CIFAR-100, and ImageNet16-120, enabling comprehensive performance comparisons. Figure \ref{STM32F7} illustrates that our hardware-aware search, incorporating performance and hardware indicators, consistently discovers highly efficient models across different latency and FLOPs constraints, with minimal performance degradation across three datasets. 
Table \ref{table2} demonstrates that our baseline results, without hardware constraints, surpass the state-of-the-art zero-shot ZiCo \cite{li2023zico}, which is achieved by optimizing condition numbers and batch size when calculating the NTK spectrum. Conversely, our proposed hardware-aware searching strategy yields a latency reduction of 1.47× to 3.23× with negligible performance trade-offs compared to TE-NAS \cite{chen2021neural}. Notably, our latency-guided search with \(\mu = 0.5/0.6\) achieves even better performance than TE-NAS \cite{chen2021neural} while reducing inference time by half on the target test MCU (Figure \ref{STM32F7}). Compared to other zero-shot methods that lack hardware awareness, our MONAS achieves better accuracy across three datasets while maintaining similar inference latency. In comparison to \(\mu\)NAS \cite{muNAS}, another NAS work targeting edge hardware, our MONAS finds models with significantly better performance and requires significantly less search time, exhibiting approximately 1104× efficiency improvement in search time (GPU hours as reported by original papers) and 6.2\% better performance when \(\mu\) is set to 0.8. Although TG-NAS \cite{qiao2024tg} has the fastest search time, it suffers from long initialization delays and lacks hardware consideration. We believe our MONAS search can be further accelerated by leveraging the proxy from TG-NAS \cite{qiao2024tg} as a performance indicator. We leave this as a research direction for further investigation.


\subsubsection{The Generalizability of MONAS}
\begin{figure}[htbp]
    \centering
    \begin{subfigure}[b]{\linewidth}
        \centering
    \includegraphics[width=\linewidth]{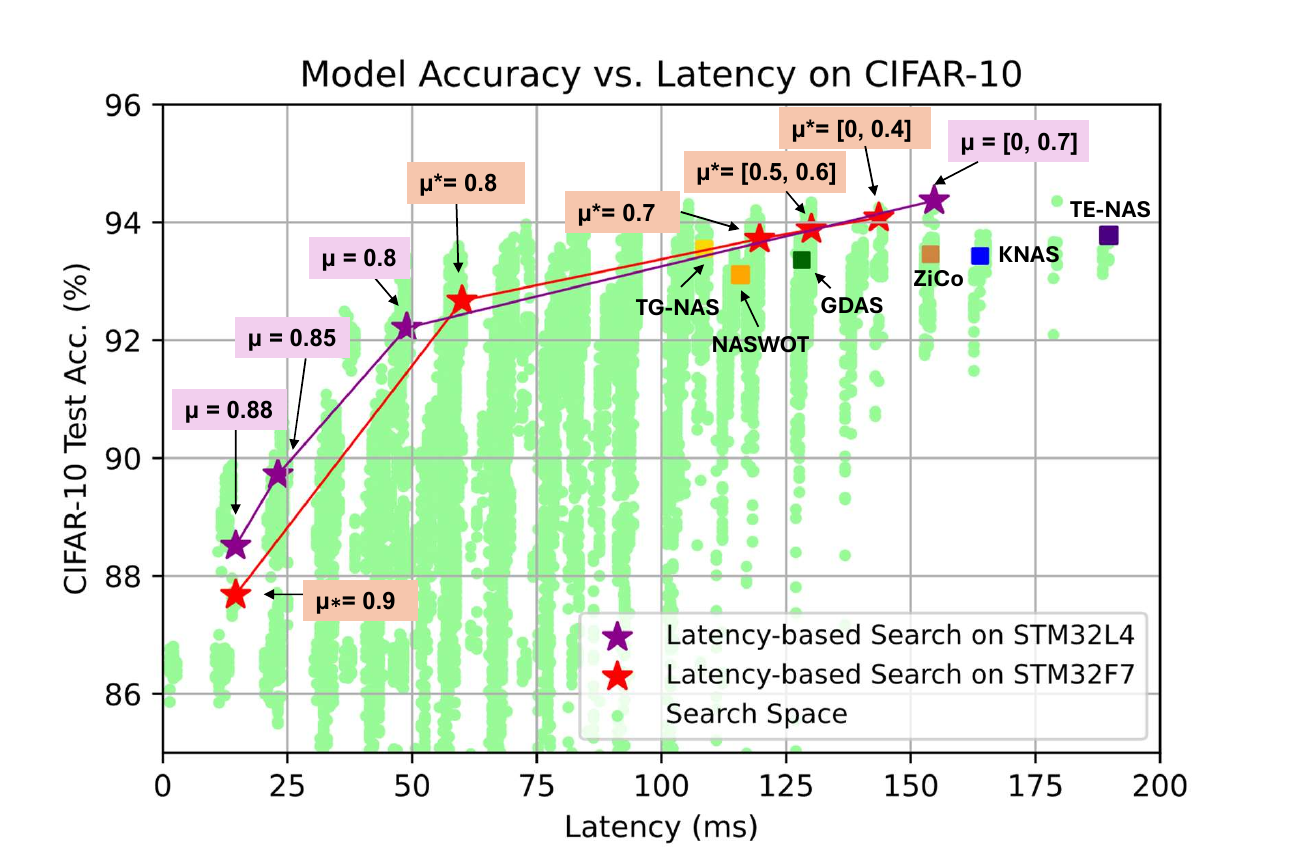}
    \end{subfigure}
    \vfill
    \begin{subfigure}[b]{\linewidth}
        \centering
    \includegraphics[width=\linewidth]{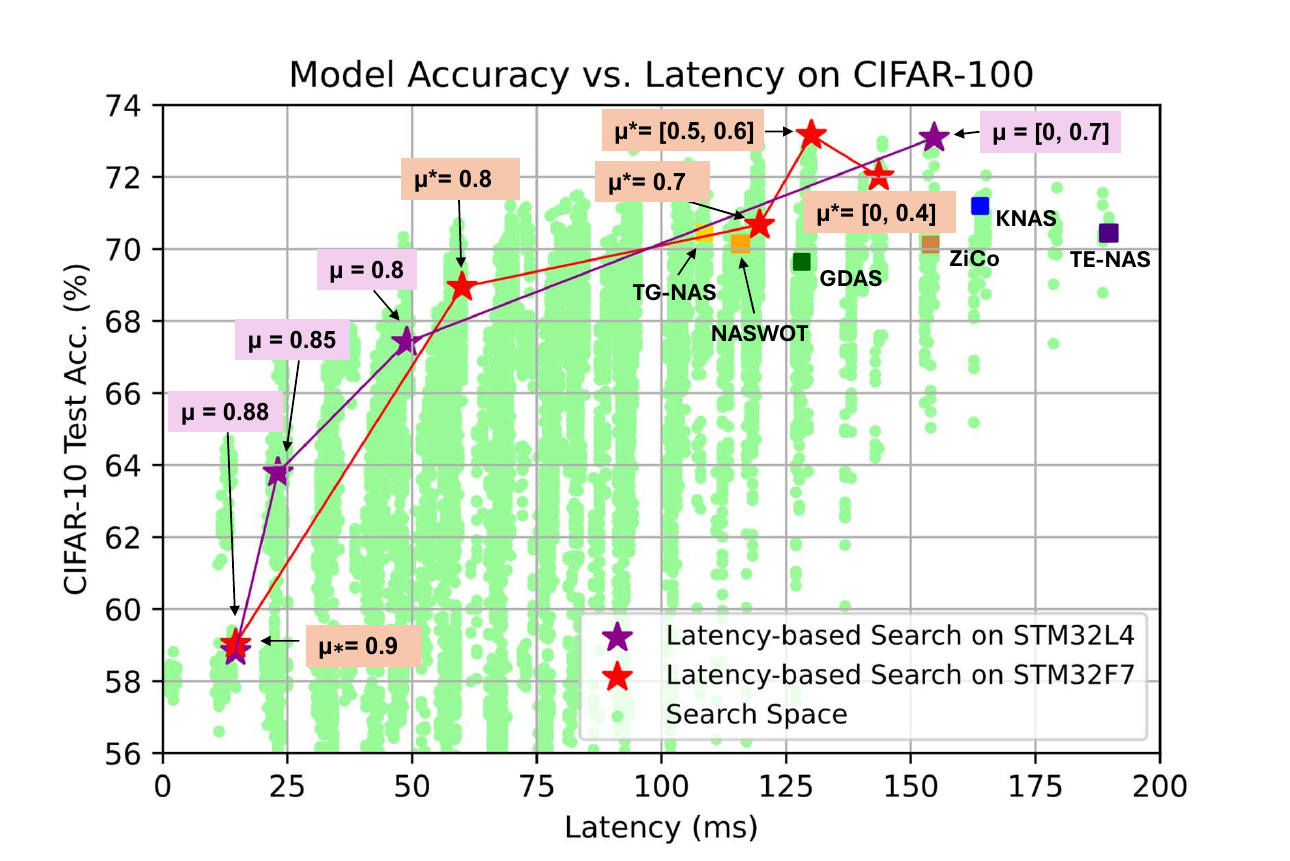}
    \end{subfigure}
    \vfill
    \begin{subfigure}[b]{\linewidth}
        \centering
    \includegraphics[width=\linewidth]{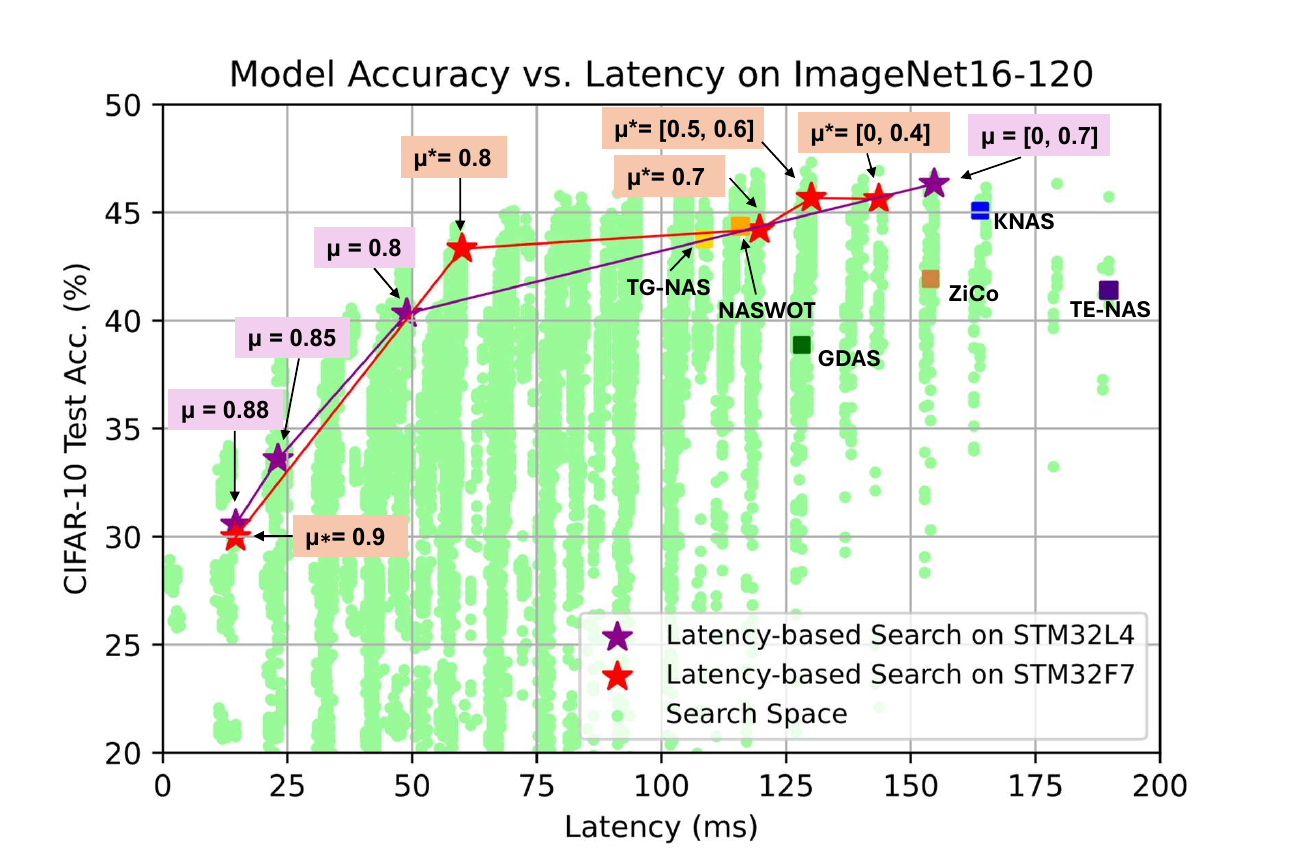}
    \end{subfigure}
\caption{Evaluation Results on STM32L4 MCU}
\label{STM32L4}
\end{figure}

To demonstrate MONAS's generalizability to other low-level devices, we further evaluate MONAS on STM32L432KC, a smaller ultra-low-power MCU with far less computation power (Cortex-M4 CPU) and memory. We first search and evaluate model architecture specifically for STM32L4, and then we evaluate the model searched with STM32F7 MCU hardware information, on STM32L4 MCU to demonstrate portability.  As Figure \ref{STM32L4} illustrates, the models searched for larger MCU (STM32F7) also perform well on smaller MCU, which shows MONAS's portability across different devices. The models specifically searched for STM32L4 perform the best, which further proves the effectiveness of MONAS. It is worth mentioning that in Table \ref{table2}, the searched model with $\mu=0.88$ has the same FLOPs and Params as $\mu=0.85$ (provided by NAS-Bench-201). However, it has a much smaller latency, which shows that MONAS can distinguish a more efficient model from a set of models with equal-value general indicators (FLOPs, Params).


\section{Conclusion}

We propose MONAS, a zero-shot neural architecture search (NAS) framework for discovering optimal deep neural network (DNN) architectures for microcontroller units (MCUs). MONAS integrates theoretical analysis of neural networks with the hardware constraints of target MCUs to identify optimal architectures with minimal accuracy degradation and negligible training and evaluation costs. Our proposed MCU latency estimation model accurately predicts real-world end-to-end inference latency and has been verified on multiple MCU platforms, with potential applicability to other edge devices. MONAS achieves a significant improvement in neural architecture search efficiency, surpassing previous MCU-targeted work by 1104×. Additionally, it discovers models that exhibit more than 3.23× faster inference on MCUs compared to generic NAS methods, while maintaining a comparable level of accuracy. We envision broader impacts of our proposed techniques when applied to other platforms and applications, and we will open source the project to facilitate future research in the field. In future work, we intend to incorporate peak memory usage modeling of MCUs to guide the search and further enhance the refinement of our MONAS framework with even faster performance indicators.






{\small
\bibliographystyle{ieee_fullname}
\bibliography{refernces}

\begin{thebibliography}{10}\itemsep=-1pt

\bibitem{abdelfattah2021zerocost}
Mohamed~S. Abdelfattah, Abhinav Mehrotra, Łukasz Dudziak, and Nicholas~D. Lane.
\newblock Zero-cost proxies for lightweight nas, 2021.

\bibitem{akhauri2024latency}
Yash Akhauri and Mohamed~S. Abdelfattah.
\newblock On latency predictors for neural architecture search, 2024.

\bibitem{cai2020onceforall}
Han Cai, Chuang Gan, Tianzhe Wang, Zhekai Zhang, and Song Han.
\newblock Once-for-all: Train one network and specialize it for efficient deployment, 2020.

\bibitem{cai2019proxylessnas}
Han Cai, Ligeng Zhu, and Song Han.
\newblock Proxylessnas: Direct neural architecture search on target task and hardware, 2019.

\bibitem{chen2021neural}
Wuyang Chen, Xinyu Gong, and Zhangyang Wang.
\newblock Neural architecture search on {ImageNet} in four {GPU} hours: A theoretically inspired perspective.
\newblock {\em arXiv preprint arXiv:2102.11535}, 2021.

\bibitem{dong2019searching}
Xuanyi Dong and Yi Yang.
\newblock Searching for a robust neural architecture in four gpu hours.
\newblock In {\em Proceedings of the IEEE/CVF Conference on Computer Vision and Pattern Recognition}, pages 1761--1770, 2019.

\bibitem{dong2020nasbench201}
Xuanyi Dong and Yi Yang.
\newblock Nas-bench-201: Extending the scope of reproducible neural architecture search.
\newblock In {\em International Conference on Learning Representations (ICLR)}, 2020.

\bibitem{he2016deep}
Kaiming He, Xiangyu Zhang, Shaoqing Ren, and Jian Sun.
\newblock Deep residual learning for image recognition.
\newblock In {\em Proceedings of the IEEE conference on computer vision and pattern recognition}, pages 770--778, 2016.

\bibitem{howard2017mobilenets}
Andrew~G Howard, Menglong Zhu, Bo Chen, Dmitry Kalenichenko, Weijun Wang, Tobias Weyand, Marco Andreetto, and Hartwig Adam.
\newblock Mobilenets: Efficient convolutional neural networks for mobile vision applications.
\newblock {\em arXiv preprint arXiv:1704.04861}, 2017.

\bibitem{jacot2020neural}
Arthur Jacot, Franck Gabriel, and Clément Hongler.
\newblock Neural tangent kernel: Convergence and generalization in neural networks, 2020.

\bibitem{kaeley2023support}
Harsimrat Kaeley, Ye Qiao, and Nader Bagherzadeh.
\newblock Support for stock trend prediction using transformers and sentiment analysis.
\newblock {\em arXiv preprint arXiv:2305.14368}, 2023.

\bibitem{kendall1938new}
Maurice~G Kendall.
\newblock A new measure of rank correlation.
\newblock {\em Biometrika}, 30(1/2):81--93, 1938.

\bibitem{krishnakumar2022nasbenchsuitezero}
Arjun Krishnakumar, Colin White, Arber Zela, Renbo Tu, Mahmoud Safari, and Frank Hutter.
\newblock Nas-bench-suite-zero: Accelerating research on zero cost proxies, 2022.

\bibitem{lee2019wide}
Jaehoon Lee, Lechao Xiao, Samuel Schoenholz, Yasaman Bahri, Roman Novak, Jascha Sohl-Dickstein, and Jeffrey Pennington.
\newblock Wide neural networks of any depth evolve as linear models under gradient descent.
\newblock {\em Advances in neural information processing systems}, 32, 2019.

\bibitem{lee2018snip}
Namhoon Lee, Thalaiyasingam Ajanthan, and Philip~HS Torr.
\newblock Snip: Single-shot network pruning based on connection sensitivity.
\newblock {\em arXiv preprint arXiv:1810.02340}, 2018.

\bibitem{lee2019snip}
Namhoon Lee, Thalaiyasingam Ajanthan, and Philip H.~S. Torr.
\newblock Snip: Single-shot network pruning based on connection sensitivity, 2019.

\bibitem{li2021hwnasbenchhardwareaware}
Chaojian Li, Zhongzhi Yu, Yonggan Fu, Yongan Zhang, Yang Zhao, Haoran You, Qixuan Yu, Yue Wang, and Yingyan Lin.
\newblock Hw-nas-bench:hardware-aware neural architecture search benchmark, 2021.

\bibitem{li2023zico}
Guihong Li, Yuedong Yang, Kartikeya Bhardwaj, and Radu Marculescu.
\newblock Zico: Zero-shot nas via inverse coefficient of variation on gradients, 2023.

\bibitem{muNAS}
Edgar Liberis, \L{}ukasz Dudziak, and Nicholas~D. Lane.
\newblock {$\mu$NAS: Constrained Neural Architecture Search for Microcontrollers}.
\newblock In {\em Proceedings of the 1st Workshop on Machine Learning and Systems}, EuroMLSys '21, 2021.

\bibitem{lin2020mcunet}
Ji Lin, Wei-Ming Chen, John Cohn, Chuang Gan, and Song Han.
\newblock {MCUNet}: Tiny deep learning on iot devices.
\newblock In {\em Annual Conference on Neural Information Processing Systems (NeurIPS)}, 2020.

\bibitem{lin2021zennas}
Ming Lin, Pichao Wang, Zhenhong Sun, Hesen Chen, Xiuyu Sun, Qi Qian, Hao Li, and Rong Jin.
\newblock Zen-nas: A zero-shot nas for high-performance deep image recognition, 2021.

\bibitem{liu2018darts}
Hanxiao Liu, Karen Simonyan, and Yiming Yang.
\newblock Darts: Differentiable architecture search.
\newblock {\em arXiv preprint arXiv:1806.09055}, 2018.

\bibitem{mei2020atomnas}
Jieru Mei, Yingwei Li, Xiaochen Lian, Xiaojie Jin, Linjie Yang, Alan Yuille, and Jianchao Yang.
\newblock Atomnas: Fine-grained end-to-end neural architecture search, 2020.

\bibitem{mellor2021neural}
Joseph Mellor, Jack Turner, Amos Storkey, and Elliot~J. Crowley.
\newblock Neural architecture search without training, 2021.

\bibitem{qiao2022two}
Ye Qiao, Mohammed Alnemari, and Nader Bagherzadeh.
\newblock A two-stage efficient 3-d cnn framework for eeg based emotion recognition.
\newblock In {\em 2022 IEEE International Conference on Industrial Technology (ICIT)}. IEEE, 2022.

\bibitem{qiao2024tg}
Ye Qiao, Haocheng Xu, and Sitao Huang.
\newblock Tg-nas: Leveraging zero-cost proxies with transformer and graph convolution networks for efficient neural architecture search.
\newblock {\em arXiv preprint arXiv:2404.00271}, 2024.

\bibitem{qiao2024micronas}
Ye Qiao, Haocheng Xu, Yifan Zhang, and Sitao Huang.
\newblock Micronas: Zero-shot neural architecture search for mcus.
\newblock In {\em 2024 Design, Automation \& Test in Europe Conference \& Exhibition (DATE)}, pages 1--2, 2024.

\bibitem{real2019regularized}
Esteban Real, Alok Aggarwal, Yanping Huang, and Quoc~V Le.
\newblock Regularized evolution for image classifier architecture search, 2019.

\bibitem{sandler2019mobilenetv2}
Mark Sandler, Andrew Howard, Menglong Zhu, Andrey Zhmoginov, and Liang-Chieh Chen.
\newblock Mobilenetv2: Inverted residuals and linear bottlenecks, 2019.

\bibitem{tan2019mnasnet}
Mingxing Tan, Bo Chen, Ruoming Pang, Vijay Vasudevan, Mark Sandler, Andrew Howard, and Quoc~V. Le.
\newblock Mnasnet: Platform-aware neural architecture search for mobile, 2019.

\bibitem{MnasNet}
Mingxing Tan, Bo Chen, Ruoming Pang, Vijay Vasudevan, Mark Sandler, Andrew Howard, and Quoc~V. Le.
\newblock {MnasNet}: Platform-aware neural architecture search for mobile.
\newblock In {\em 2019 IEEE/CVF Conference on Computer Vision and Pattern Recognition (CVPR)}, pages 2815--2823, 2019.

\bibitem{wang2020picking}
Chaoqi Wang, Guodong Zhang, and Roger Grosse.
\newblock Picking winning tickets before training by preserving gradient flow, 2020.

\bibitem{white2023neural}
Colin White, Mahmoud Safari, Rhea Sukthanker, Binxin Ru, Thomas Elsken, Arber Zela, Debadeepta Dey, and Frank Hutter.
\newblock Neural architecture search: Insights from 1000 papers, 2023.

\bibitem{xiao2020disentangling}
Lechao Xiao, Jeffrey Pennington, and Samuel Schoenholz.
\newblock Disentangling trainability and generalization in deep neural networks.
\newblock In {\em International Conference on Machine Learning}, pages 10462--10472. PMLR, 2020.

\bibitem{xiong2020number}
H. Xiong, L. Huang, M. Yu, L. Liu, F. Zhu, and L. Shao.
\newblock On the number of linear regions of convolutional neural networks, 2020.

\bibitem{knas}
Jingjing Xu, Liang Zhao, Junyang Lin, Rundong Gao, Xu Sun, and Hongxia Yang.
\newblock {KNAS}: Green neural architecture search.
\newblock In {\em Proceedings of ICML 2021}, 2021.

\bibitem{yang2020scaling}
Greg Yang.
\newblock Scaling limits of wide neural networks with weight sharing: Gaussian process behavior, gradient independence, and neural tangent kernel derivation, 2020.

\bibitem{ying2019nasbench101}
Chris Ying, Aaron Klein, Esteban Real, Eric Christiansen, Kevin Murphy, and Frank Hutter.
\newblock Nas-bench-101: Towards reproducible neural architecture search, 2019.

\bibitem{zela2022surrogate}
Arber Zela, Julien Siems, Lucas Zimmer, Jovita Lukasik, Margret Keuper, and Frank Hutter.
\newblock Surrogate nas benchmarks: Going beyond the limited search spaces of tabular nas benchmarks, 2022.

\end{thebibliography}
}

\end{document}